\documentclass[runningheads]{llncs}
\usepackage{graphicx}
\usepackage{forloop}
\newcounter{loopc}

\usepackage{todonotes}
\usepackage{nicefrac}

\begin{document}
%
\title{Limitations of Assessing Active Learning Performance at Runtime}
%
\titlerunning{Limitations of Assessing Active Learning Performance at Runtime}
%

\author{Daniel Kottke\inst{1}\orcidID{0000-0002-7870-6033} \and
Jim Schellinger\inst{1} \and \\
Denis Huseljic\inst{1}\and
Bernhard Sick\inst{1}}
\authorrunning{D. Kottke et al.}
%
\institute{Intelligent Embedded Systems, University of Kassel, Kassel, Germany 
\email{\{daniel.kottke,bsick\}@uni-kassel.de}\\
\email{\{jschellinger,dhuseljic\}@student.uni-kassel.de}\\
\url{http://www.ies-reserach.de} }
\maketitle              
\begin{abstract}
Classification algorithms aim to predict an unknown label (e.g., a quality class) for a new instance (e.g., a product). Therefore, training samples (instances and labels) are used to deduct classification hypotheses. Often, it is relatively easy to capture instances but the acquisition of the corresponding labels remain difficult or expensive.
Active learning algorithms select the most beneficial instances to be labeled to reduce cost. In research, this labeling procedure is simulated and therefore a ground truth is available. But during deployment, active learning is a one-shot problem and an evaluation set is not available. Hence, it is not possible to reliably estimate the performance of the classification system during learning and it is difficult to decide when the system fulfills the quality requirements (stopping criteria).
In this article, we formalize the task and review existing strategies to assess the performance of an actively trained classifier during training. Furthermore, we identified three major challenges: 1)~to derive a performance distribution, 2)~to preserve representativeness of the labeled subset, and 3) to correct against sampling bias induced by an intelligent selection strategy. In a qualitative analysis, we evaluate different existing approaches and show that none of them reliably estimates active learning performance stating a major challenge for future research for such systems. All plots and experiments are provided in a Jupyter notebook that is available for download.


\end{abstract}
\section{Introduction}

Machine learning algorithms recently have been remarkably successful in applications where lots of data is available. This article focuses on problems where unlabeled data is easily available but acquiring the corresponding labels might be expensive because human experts or massive calculations are necessary. Typical examples are credit scoring~\cite{Krempl2017}, medical mining~\cite{Jothi2015}, or quality inspection (QI)~\cite{Douak2012}. QI systems, for example, use data from various sensors (e.g., cameras, pressure sensors, etc.) to build a classifier that is able to detect faults in produced items. Features can easily be yielded by these sensors but labels (e.g., quality classes) need further investigation by experts or require more tests which might take a long time or induce further cost. The field of active learning (AL) investigates which item should be selected to be manually labeled in order to improve the automatic classifier most effectively. Here, remarkable results have been achieved in research environments~\cite{Settles:2009} where label acquisition is simulated and the ground truth is known~\cite{KCH+17}.

However, in a real-world setting, the AL task is a one-shot problem and the ground truth is not known. 
The major challenge in realistic AL scenarios (one-shot problems) is to \emph{estimate the performance of the iteratively trained classifier at runtime}. 
Without that estimate, a human operator is not able to supervise the costs induced by the active classification system. These costs are induced by the oracle (labeling cost) and by wrong decisions of the classifier (misclassification cost). Without a reliable performance estimate the misclassification cost cannot be estimated and a stopping criterion for real-world applications cannot be found.
To our knowledge, such a performance estimate has never been explicitly studied in the context of AL.

This article summarizes different approaches that might qualify as an performance estimate. These approaches are either derived from the idea of a selection strategy or from related areas of AL. Thereby, we identified three challenges:
 1)~In AL, performance estimators are faced with the highly volatile nature of tasks with limited supervision. Hence, a single performance estimate might be misleading. A stochastic model of performance (performance distribution) could ensure reliability for performance estimation.
 2)~Assessing reliable performance need hold-out data for evaluation. Nevertheless, sampling additional labels is counterproductive. Also, holding back sampled data for evaluation reduces the sampled information. Hence, a representative training set is necessary to assess the real performance.
 3)~Selecting only the most useful instances in AL indicates that the sampling distribution will be different from the original data-generating distribution. This sampling bias has an effect on evaluation and should be considered.
Our experimental evaluation shows that non of these algorithms are able to reliably estimate the performance -- even on a very simple dataset.
%
%
The main contributions of this article are:\\~\vspace{-1.8em} 
\begin{itemize}
  \item We define the problem of assessing the classifier's performance during AL and propose two baseline strategies to evaluate performance estimation.
  \item We provide a comprehensive survey of related work, discuss three major challenges, and identify approaches to estimate performance at runtime.
  \item We evaluate the approaches and provide all figures and experiments in a Jupyter notebook\footnote{Download the Jupyter notebook at: \url{http://www.ies.uni-kassel.de/p/perfest}} for reproducible research.
  \item Our results show the demand for more advanced performance estimators.
\end{itemize}



The article starts with a survey on performance estimation in active learning, followed by a presentation of the most prominent challenges. Next, we identify approaches to assess performance at runtime which are then evaluated. The article finishes with a conclusion and a discussion of future work.

\section{Background and Related Work}

\begin{figure}
  \centering
  \includegraphics[width=.7\textwidth]{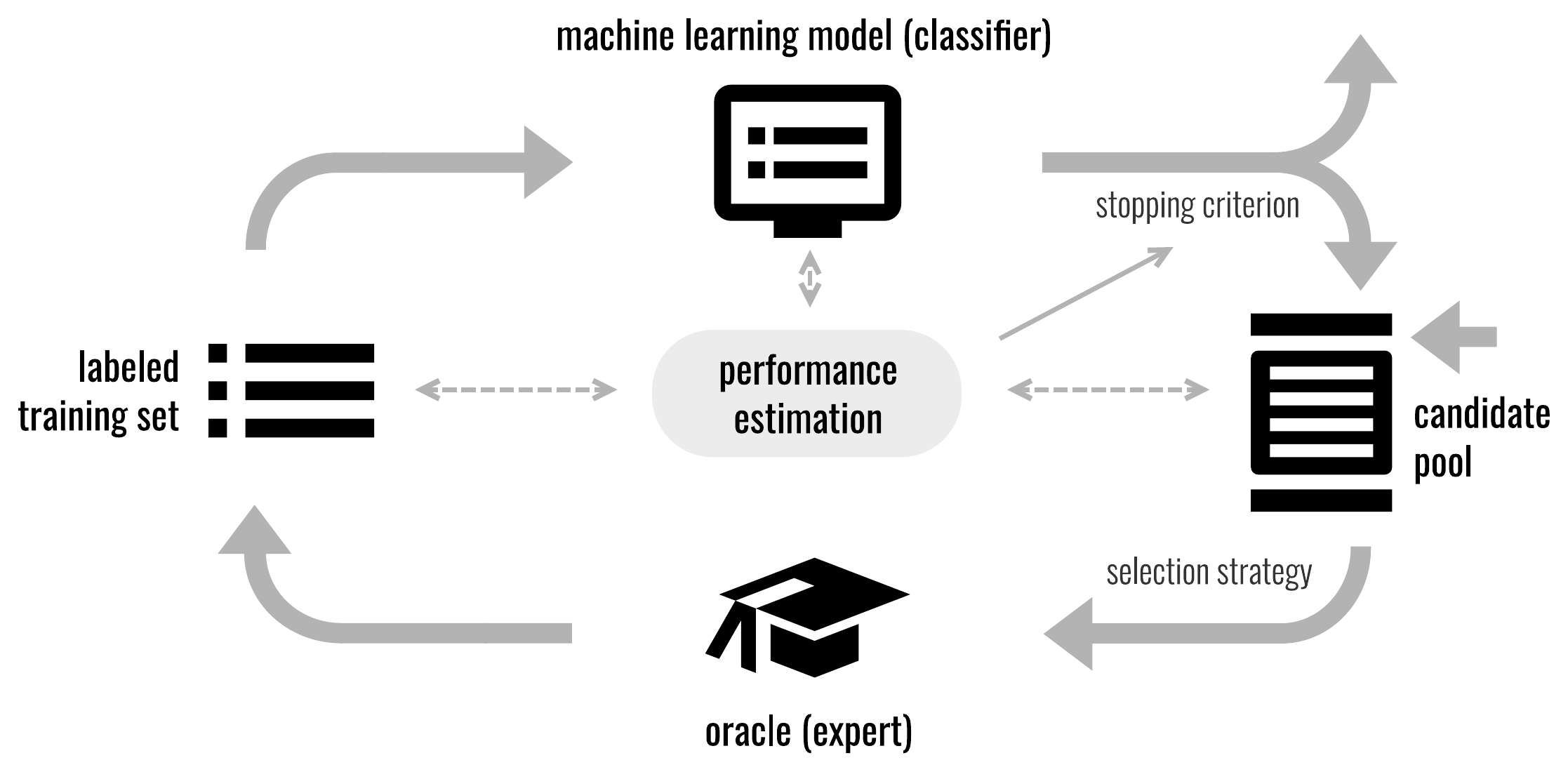}
  \caption{Active learning cycle~\cite{KCH+17} showing the additional performance estimation component using information from the classifier, the labeled set and the candidate pool.} \label{fig:al_cycle}
\end{figure}

In active learning (AL), we assume to have a fully unlabeled (or partially labeled) dataset at the beginning. Successively, labels are acquired from an oracle to build the training basis for the classifier~(see Fig.~\ref{fig:al_cycle}). The goal is to select those instances for labeling that improve the classifier the most according to an application-specific performance measure. Thereby, the number of labels should be reduced, which saves cost or time~\cite{Settles:2009}.

The central question in AL is: How to select the next instance? The idea is to quantify the usefulness of every labeling candidate and select the most promising one. In this article, we focus on estimating the performance of the iteratively trained classifier. Therefore, information of the classifier itself, the labeled subset and the candidate pool might be required~(see Fig.~\ref{fig:al_cycle}). This estimate is essential to define a well-suited stopping criterion. 

In this section, we summarize algorithms that might help assessing the performance at runtime. Often the aspect of performance estimation was only implicitly mentioned.

\subsection{Measuring the Usefulness: Selection Strategies}
As the estimation of performance in AL has not been intensively researched yet, we first concentrate on the development of selection strategies as these aim to optimize the performance of the classifier. 

One of the most commonly used selection strategy is \emph{uncertainty sampling}~\cite{LG94}. The idea is that labels are preferably acquired where the classifier is most uncertain (e.g., near the decision boundary). Thereby, it aims at resolving uncertain class assignments. Depending on the classifier and the problem's difficulty, this might lead to a self-locked in problem~\cite{Dasgupta2011}, i.e., some regions in the feature space are never explored as the result is biased by the technique. This might lead to poor results  although the classifier seems to be very confident~\cite{conf/ecai/KottkeKLTS16}. 

\emph{Query by Committee}~\cite{Seung:1992}  trains a classifier ensemble and use the disagreement between classifiers as a proxy to describe the usefulness of a label acquisition. Sampling focuses on reducing variance between the classifiers (version space partitioning~\cite{Settles:2009}) and aims at stabilizing classifier predictions. 

A decision-theoretic approach, called \emph{Expected Error Reduction}~\cite{Roy01towardoptimal}, determines the classification error on the current labeled set and on all possible future labeled sets if one additional label would be acquired. The error score is measured by the so-called generalization error using a set of instances (without the need for labels) and the posterior probabilities using Eq.~\ref{eq:generalization_error} with $x_i \in X$ being the instances, $Y = \{1, \dots, C\}$ being the classes ($C$ is the number of classes), and $\mathcal{E}$ being the indices of instances used for evaluating the performance.
\begin{equation} \label{eq:generalization_error}
  err = \sum_{i \in \mathcal{E}} 1-\max_{y \in Y} \hat{p}(y \mid \vec{x}_i)
\end{equation}
Later, Chapelle~\cite{chapelle2005active} recognized that these posterior probabilities are highly unstable when only a few labels are available. He proposed to use a beta-prior which simulates that each class has one global member with weight $\epsilon$. Hence, the posterior probabilities in totally unlabeled regions stay nearly uniform.

In \emph{Importance Weighted AL}, Beygelzimzer~et~al.~\cite{Beygelzimer2008a} claim to correct the bias induced by sampling. The  unlabeled instances are processed one-by-one in a data stream and a coin is flipped with probability $p_t$ to acquire a label or not. If a label is acquired, the algorithm weights this instance with $\nicefrac{1}{p_t}$  correcting for sampling bias. The rejection-threshold $p_t$ is determined by a method called loss-weighting. The extension \emph{Agnostic Active Learning}~\cite{Beygelzimer2010} works without keeping the version space. 
\emph{Probabilistic AL}~\cite{krempl2015optimised} combines the expected error reduction approach and the idea of uncertainty sampling in one single decision-theoretic approach. Instead of evaluating the performance increment on a full evaluation set (like expected error reduction), Krempl~et~al. use local label statistics to derive a distribution of the true posterior probability using a Beta distribution. Then, they use the idea of expected error reduction to determine the gain in performance by simulating future label acquisitions in that local region.
Reitmeier~et~al.~\cite{REITMAIER2013106} proposed \emph{4DS} which considers the distance of samples to the decision boundary, the density in regions, where samples are selected, the diversity of samples in the query set, and the unknown class distribution.

\subsection{Stopping Criterion}
Performance estimates are necessary in order to decide when to stop the acquisition of labels. This subsection summarizes the most common approaches as in many studies this problem is not considered at all. Laws~et~al.~\cite{Laws2008} presented three optimization tasks to stop learning: \emph{1)~Minimal Absolute Performance} stops learning when a certain user-defined threshold is reached. Therefore the classifier must be able to estimate its own performance (e.g., accuracy, f-score). \emph{2)~Maximum Possible Performance} stops learning when the performance can no longer be improved. \emph{3)~Convergence} stops when additional samples from the pool does not contribute more information.  
For some learning models, this could be quite similar to (2).
The choice of optimization functions depends on the application and the user's demands. Similar to \cite{Wang2014}, we categorized the methods in i)~performance~based, ii)~gradient~based, and iii)~confidence~based.

{Performance-Based:}
Laws~et~al.~\cite{Laws2008} estimate the f-score of a classifier in an AL process by calculating the true positive rate, true negative rate, and false negative rate extending the idea from the generalization error used in expected error reduction. Predicting the f-score in each training iteration, they stop acquiring labels as soon as a desired f-score value is reached.
A meta-learning approach is used by Naik~et~al.~\cite{Naik2013}. They create features from the classification model at certain time points and train a regression algorithm to learn a model-performance-mapping. Using this mapping, they predict the performance.

Gradient-Based: Wang~et~al.~\cite{Wang2014} compute the gradient of the empirical error for a given algorithm and stop if all remaining samples in the pool barely change the model. In \cite{Wang2014}, they apply their methods to Logistic Regression and Support Vector Machines.
Laws~et~al.~\cite{Laws2008} compute the gradient of either performance (based on the generalization error) or uncertainty (based on class posteriors) to determine whether the point is reached, where the pool has become uninformative and no sample will significantly change the model as the AL performance converged.

Confidence-Based:
For confidence-based stopping criteria, a user-defined uncertainty value is assumed as a good measure to determine whether a classifier is confident or not. Observing only small uncertainty values, it is assumed that the classifier is confident.
Vlachos~\cite{Vlachos2008} proposed a method for support vector machines, maximum entropy models and Bayesian logistic regression. After each acquisition, he computes the confidence (i.e., the average margin of a SVM or the entropy in Bayesian logistic regression). Once the confidence starts dropping, the AL process is stopped because no informative samples are left in the pool.
Zhu~et~al.~\cite{Zhu2010} give an introduction to the confidence-based stopping criterion. Furthermore, they propose methods on how to stop the AL algorithm. Their \emph{maximum uncertainty method} stops, if all samples from the unlabeled pool have an uncertainty value below a user-defined threshold. Additionally, they describe an \emph{overall uncertainty method} that uses the average uncertainty of the pool.

Others~\cite{Olsson2009,Bloodgood2009b} use a committee of classifiers and argue that the agreement indicates confidence. They define different agreement measures that can be derived from the unlabeled pool or from the labeled training set. Olsson et~al.~\cite{Olsson2009} propose to use an intrinsic stopping criterion that consists of the \emph{selection agreement}~\cite{Tomanek2007} and \emph{validation set agreement}~\cite{Tomanek2008}. Bloodgood~\cite{Bloodgood2009b} uses the Kappa statistic~\cite{Cohen1960} agreement measure and proposed a method that provides stopping adjustability.The algorithm allows to stop earlier to save human annotation effort or to stop later to increase the chance of reaching the optimal performance.


\subsection{Unsupervised Risk Estimation and Related Topics}
In 2010, Domnez~et~al.~\cite{Donmez2010} formalized the problem of unsupervised risk estimation. Multiple independent risk estimators are used to be applied to any arbitrary dataset.Platanios~et~al.~\cite{Platanios2014}, uses the concept of competing boolean classification functions, to estimate the accuracies as well as error dependencies throughout their agreement rates on unlabeled data. 
The method was enhanced by probabilistic logical class constraints~\cite{Platanios2017} and a Bayesian approach inferring the posterior distribution~\cite{Platanios2016}. However, \cite{Platanios2014} and~\cite{Platanios2017} are limited by using multiple boolean classifiers under the assumption of independent error.
Jaffe~et~al.~\cite{Jaffe2015} derives an algorithm to estimate the class imbalance and calculate the true positive and true negative rate under the assumption of independent errors between a given set of classifiers.
Biltzer~et~al.~\cite{Blitzer2011} resolves the concept of domain adaption, using unlabeled data to extract weights, linking features derived through the training data set to a divergent target data.
Steinhardt~et~al.~\cite{Steinhardt2016} claim to soften the assumptions by using a three-view approach. This means, instead of modeling conditional distributions which require certain parametric families, they split the feature vector $\vec{x}$ into three conditional independent (given the class) parts. Using the different loss values for each class, they are able to determine the expected risk. Unfortunately, these rather theoretic approaches require intensive research to be applied to performance estimation in AL.

Liu~et~al.~\cite{LiuJunGhosh2009} mentioned that sampling labels according to uncertainty sampling states some issues for cost-sensitive applications. Their approach is to train a classifier on the labeled training set and predict the class labels for the complete candidate pool. Then, the final cost-sensitive classifier is trained on both the labeled training set and the self-labeled candidate pool.
The term sampling bias has been defined more formally in~\cite{Dasgupta2011}: During training, the sampling distribution differs more and more from the data generating distribution as solely the most informative labels are acquired. The authors mention that "sampling bias is the most fundamental challenge posed by active learning".

\section{Challenges of Performance Estimation in AL}

Facing real applications, active learning (AL) is a one-shot problem. This means that additional labels for evaluation are not available and should not be acquired. This would be counterproductive as it increases the cost of the labeling process that we aim to minimize. In this section, we summarize the most prominent challenges when estimating the performance at runtime in a one-shot setting. Furthermore, we provide illustrative examples for visualization.

\subsection{Objective Function: Which performance is to be estimated?}
In the first setting, the classifier must be evaluated on a separate hold-out dataset, whereas in the second case the classifier would be evaluated on the predicted and the acquired labels~\cite{Tong2001}.

We propose to use two different baselines for experimenting with performance estimation in AL. Given a set of labeled instances and knowledge about the data-generating distribution, the \textbf{true baseline} describes the performance yielded on an ``infinitely'' large evaluation set that is sampled from the data-generating distribution. Given a classification model trained on the labeled set, this estimate is deterministic. Nevertheless, this score could never be achieved in an AL task as we do not know the full data-generating distribution. Hence, we are not able to build an ``infinitely'' large evaluation set. In contrast, having labeled only $B$ instances, we maximally are able to evaluate on $B$ instances. Accordingly, the \textbf{subsample baseline} samples $B$ labeled instances from the data-generating distribution multiple times and determines a distribution of performance. Note that the expectation value of infinite subsample baselines equals the true baseline.


\subsection{Performance Distribution}

\begin{figure}[b]
  \centering
  \includegraphics[width=.6\textwidth]{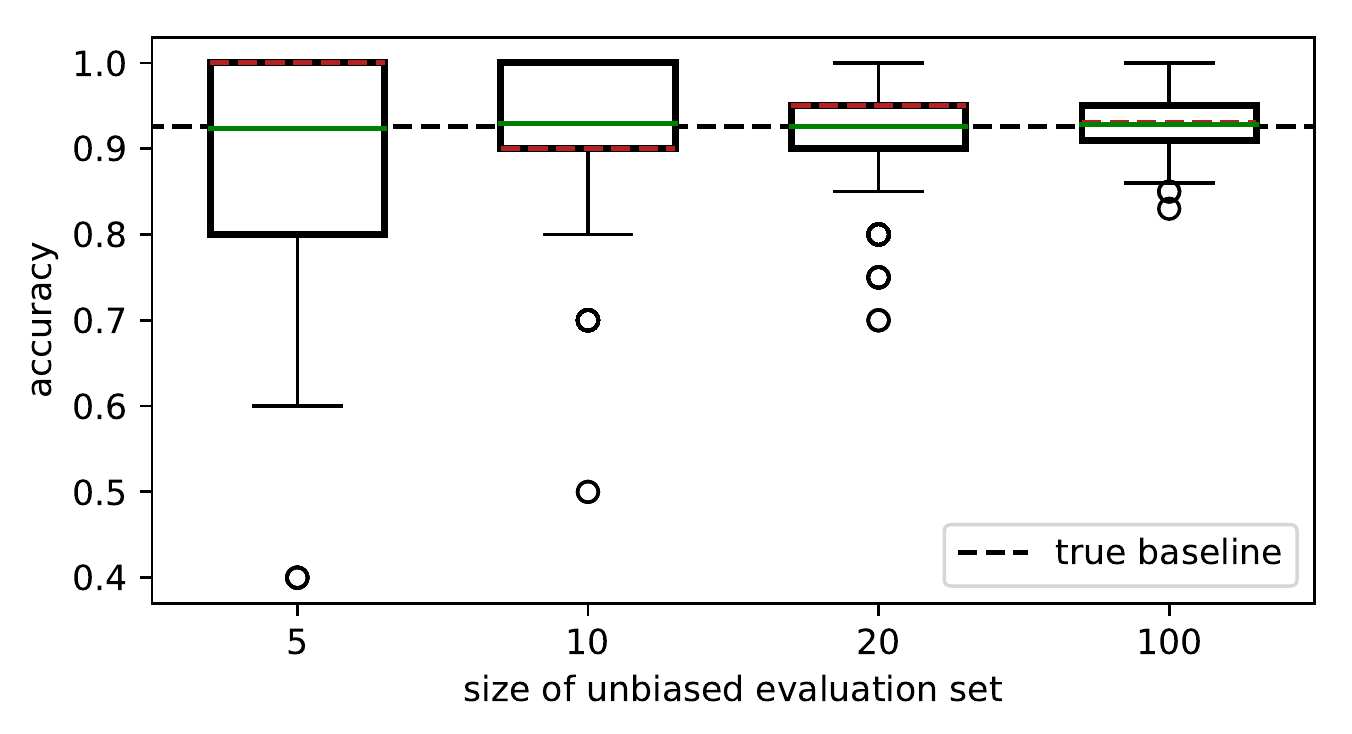}
  \caption{Accuracy distributions with differently large evaluation sets tested on a fixed classifier. The reliability of a single performance value might be misleading if the evaluation set is too small. We used boxplots with mean in green and median in red.}\label{fig:performance_distribution}
\end{figure}

\begin{figure}[b!]
  \centering
  \includegraphics[width=.65\textwidth]{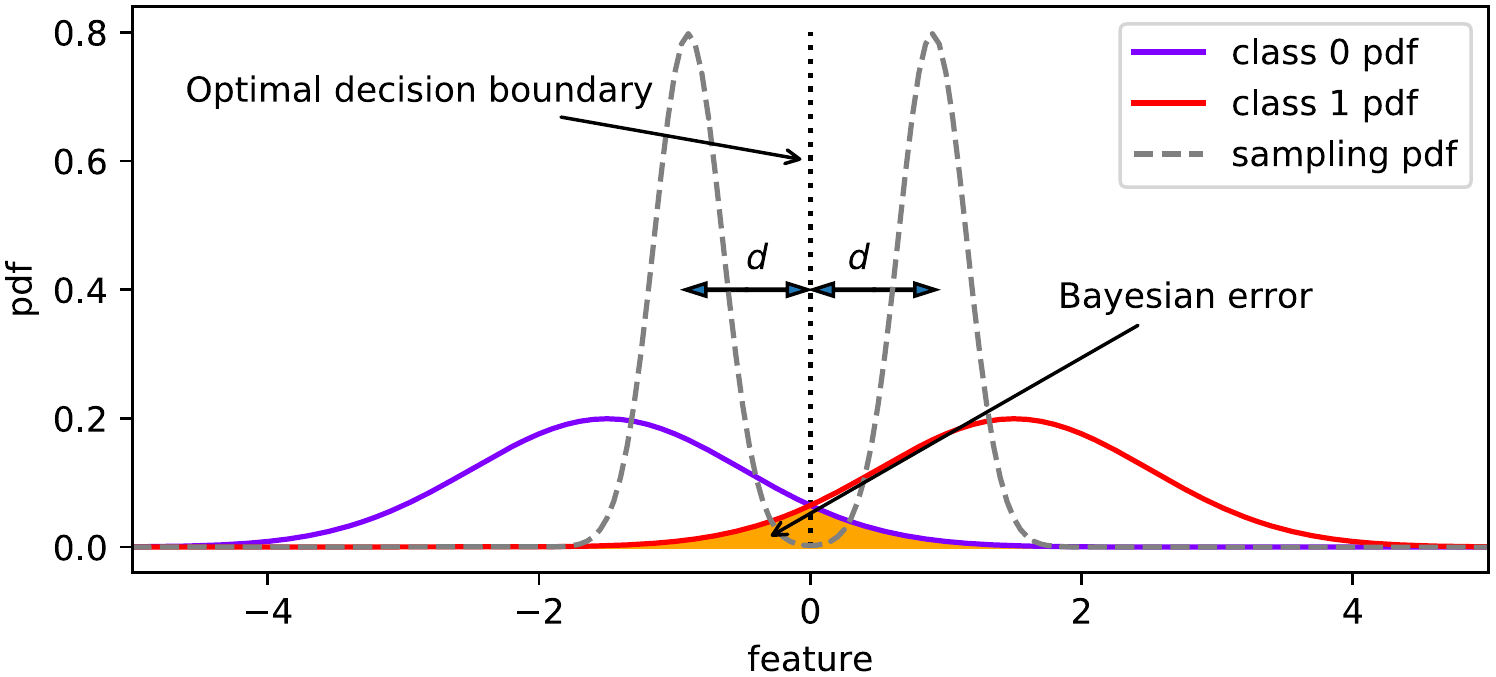}
  \caption{Class-conditional probability distribution function (pdf) of our learning task (in color). The optimal decision boundary is at $0$. The Bayesian error area is shown in orange. The dashed pdf shows an exemplary sampling distribution for $d=0.9$.}\label{fig:samplingbias_dataset}
\end{figure}

Having available only a certain number of randomly sampled evaluation instances emphasizes the necessity to describe this random process.
Fig.~\ref{fig:performance_distribution} shows the performance distribution of a fixed Parzen window classifier~\cite{chapelle2005active} evaluated on multiple (1000 times) unbiased evaluation sets (randomly sampled) with different sizes ($5, 10, 20, 100$). 
The data consists of two classes with equal class prior. A standard normal distribution with means $-1.5$ (class 1) and $1.5$ (class 2), respectively, (see Fig.~\ref{fig:samplingbias_dataset}) is used to model the data. 
The true baseline was evaluated on an independent set large enough to create stable results (here: 2000 samples).
The classifier was trained on 100 randomly selected instances. 

%

Estimating a single accuracy value of a classifier with a small number of evaluation instances, the reliability of this value might be questionable. In Fig.~\ref{fig:performance_distribution}, a small evaluation subset is not stable enough (high variance) even for such an easy classification problem. Evaluating on only 5 instances, in half of the cases the estimate will provide accuracy $100\%$ ($5$ out of $5$ predictions are correct) or below $80\% $. An over or under estimation of our classifier is much more likely if no further evaluation samples are acquired. Note that this study is executed on a very easy example for illustration purposes. Real world applications would even require much larger evaluation sets to achieve reliable (within a certain confidence interval) results. 

\subsection{Representativeness of the Training Set in AL Evaluation} 
In AL, using some of the labels for evaluation and thereby reducing the number of labels for training is counterproductive as this biases our performance estimate compared to a classifier trained on the complete set. 


\begin{figure}[t!]
  \centering
  \includegraphics[width=.6\textwidth]{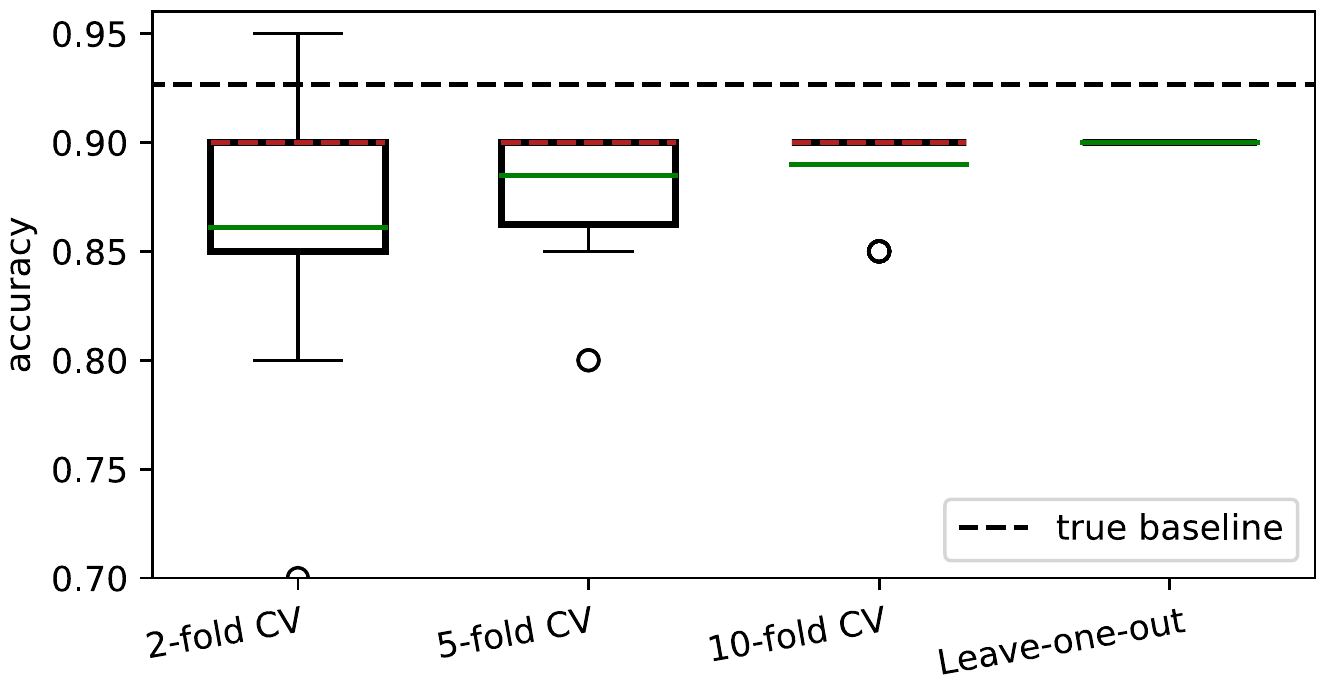}
  \caption{Accuracy estimation distributions for different cross-validation (CV) approaches. Depending on the number of folds, the size of the training set changes. This induces different biases to the true performance (determined with all acquired labels).}\label{fig:representativeness}
\end{figure}

Fig.~\ref{fig:representativeness} shows the effect of reducing the training set to save labels for evaluation. We acquired 20 labels and conducted a 2-, 5-, 10-fold. and a leave-one-out cross-validation. In each of 50 repetitions, the mapping of an instance to a fold is randomized. Hence, results may vary across repetitions.

The figure shows that the estimation accuracy -- difference of the mean (green line) and the true baseline -- improves as the number of training instances increases. For example, performing a 2-fold cross-validation with 20 labeled instances, we have $2$ training sets with $10$ instances each. Performing a leave-one-out cross-validation, the training set size is $19$ instances. Hence, the most accurate estimation can be achieved by using as many labels as possible for training. As there is only one way to split $20$ instances into $20$ folds, we do not get information about the variance induced by the evaluation methodology.

\subsection{The Influence of Sampling Bias}

Selection strategies in AL aim to select the most informative instances. Hence, the instance space is specifically sampled according to the classifiers characteristics and the performance function to be optimized. As a consequence, the sampling distribution is no longer similar to the original data-generating distribution. Hence, this sample is not appropriate for evaluation purposes. 

\begin{figure}[b!]
  \centering
  \includegraphics[width=.6\textwidth]{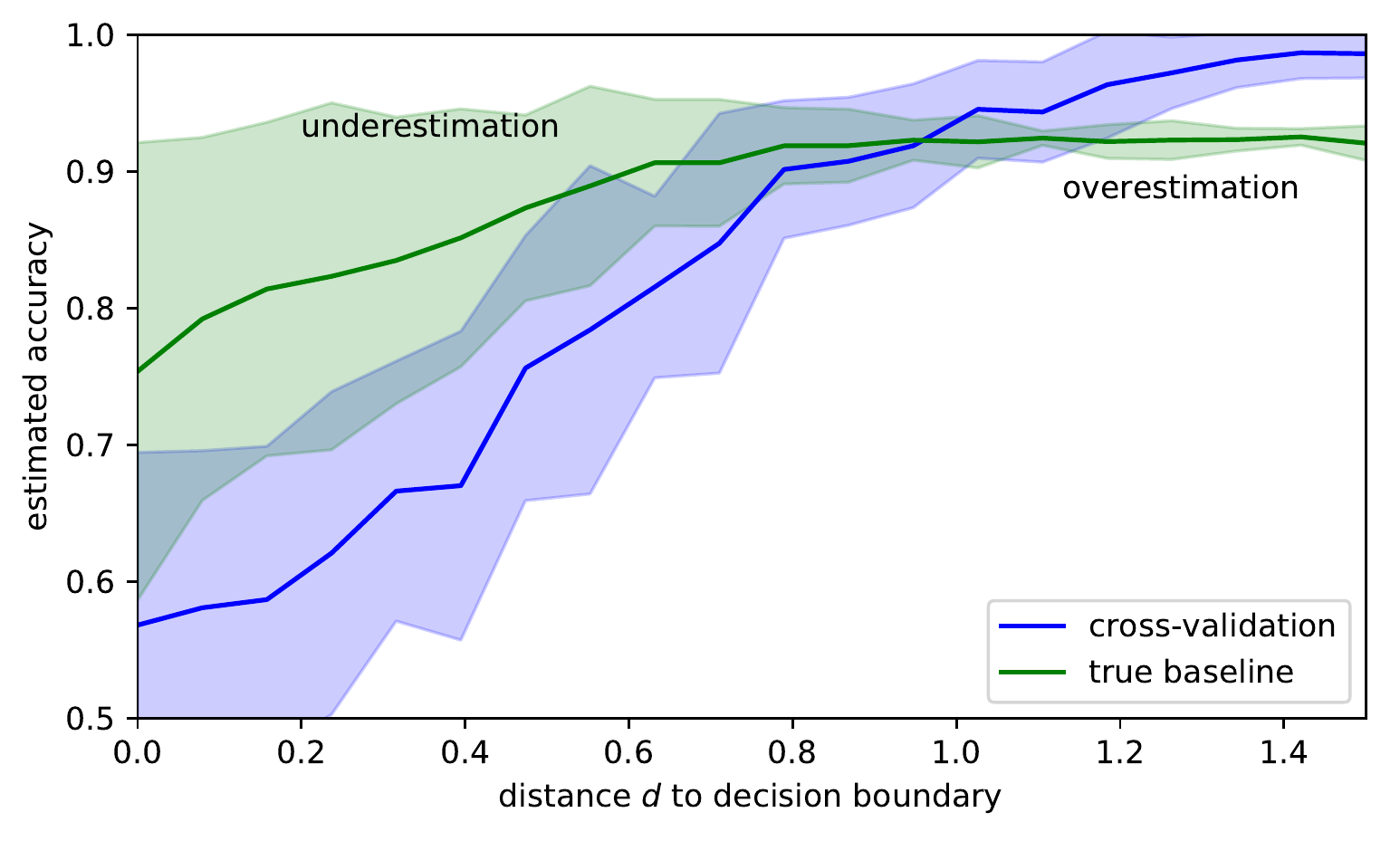}
  \caption{Performances of internal cross-validation and the true baseline performance. Depending on how the samples are biased by the selection strategy, the performance estimate might be either optimistic or too pessimistic.}\label{fig:samplingbias_performance}
\end{figure}

We show the effect of sampling bias on the performance estimate using the dataset described before (see Fig.~\ref{fig:samplingbias_dataset}). To identify the effect in a controlled environment, we chose to define a sampling distribution instead of using some selection strategy. To show the influence of the sampling bias, the sampling distribution consists of two normal distributions ($\mu = \pm d, \sigma=\nicefrac{1}{4}$) with equal prior and a specified distance $d$ from the optimal decision boundary. The data-generating distribution remain fixed.

In Fig.~\ref{fig:samplingbias_performance}, we evaluate the labeled set (containing $30$ instances) using a 3-fold cross-validation. Hence, the training set is build using 2 folds. We evaluate the classifier on the remaining fold to determine the cross validation performance and on a separate hold-out evaluation set to determine the true baseline. The plot shows means and standard deviations from $50$ repetitions.

Sampling in the area of the Bayesian error (small $d$) can be compared to selection strategies that acquire labels in regions with high uncertainty, such as uncertainty sampling. Here, only few samples were selected in unambiguous regions. When assessing the classifier's performance (e.g., using cross-validation), it is trained and evaluated on difficult instances, thus the accuracy derived by internal cross-validation strongly underestimates the true classifier performance as easy instances (that are sufficiently present in the original task) are rarely considered. Sampling distant to the decision boundary (large $d$), we ignore the Bayesian error and train and evaluate only on easy instances. We tend to overestimate the classifiers performance. Note that for more complex problems, sampling far from the decision boundary may strongly decrease the true performance. 

%

\newpage
\section{Assessing Classifier Performance at Runtime}
\label{sec:approaches}

In this section, we describe different approaches to assess the classifier's performance at runtime without acquiring additional labels. These ideas are extracted from our study of related work and standard performance assessment approaches. We evaluate the algorithms and the proposed baselines on our previously mentioned dataset (see Fig.~\ref{fig:samplingbias_dataset}). The results show that none of the mentioned algorithms is able to reliably estimate the performance during runtime although this dataset is very simple. Hence, we did not extend the study for multiple datasets.
Using the published Jupyter notebook code and visualizations, one may also apply a certain dataset if interested. 

\subsection{Algorithms}
This \textbf{Generalization Error} has been used to measure the expected error in the selection strategy of Roy and McCallum~\cite{Roy01towardoptimal}. The idea is similar to the true positive (resp. negative) rate proposed by Laws~et~al.~\cite{Laws2008} to define a performance-based stopping criterion. It is calculated according to Eq.~\ref{eq:generalization_error}. For comparability purposes, we present the accuracy performance calculated by $\mathrm{accuracy} = 1-\mathrm{error}$.
The \textbf{k-fold Cross-validation} (here: $k=3$) is one of the most commonly used evaluation principles. The idea is to create $k$ folds of the training set. Then one of these sets provide evaluation instances whereas the instances from the other folds are used for training the classifier~\cite{Bishop2006}. Evaluating all folds, we get independent predictions for all labeled instances. Then, the accuracy of the classifier is the relative frequency of correct predictions.
The \textbf{Self-Labeling With Cross-Validation} approach is motivated by Liu~et~al.~\cite{LiuJunGhosh2009} who propose to use self-labeling, i.e., a classifier provides labels for the candidate pool. Then, a new classifier is trained on both the labeled set and the self-labeled candidate set. In our performance estimation task, we use this combined set to perform a cross-validation. Similar to above, we use 3 folds but only a random selection of instances such that the number of training instances is equal to the size of the originally labeled set.
In \textbf{Reweighted Cross-Validation}, we use the idea of importance weighted and agnostic AL~\cite{Beygelzimer2008a,Beygelzimer2010} which is to weight the instances according to the inverse of their sampling probability. Hence, reweighted cross-validation describes a standard cross-validation which weights the predictions according to the inverse of the corresponding instance's sampling probability.
The \textbf{Probabilistic Performance} follows the idea of Krempl~et~al.~\cite{krempl2015optimised} who proposed to model the true posterior with label statistics (the estimated posterior probability for a positive class $\hat{p}$, and the number of nearby labels $n$) using a Beta distribution. Furthermore, they showed how to derive the local (for a given instance) accuracy distribution using this model. The accuracy distribution for one instance is a Beta distribution with parameters $\alpha=1+\max(n \hat{p}, n (1-\hat{p}))$ and  $\beta=1+\min(n \hat{p}, n (1-\hat{p}))$.
Having an unbiased evaluation set of samples (labeled or unlabeled), we can calculate the mean probabilistic performance using a mixture of Beta distributions with equal priors.

\subsection{Experimental Design}
\begin{figure}[p]
  \centering
  \includegraphics[width=.92\textwidth]{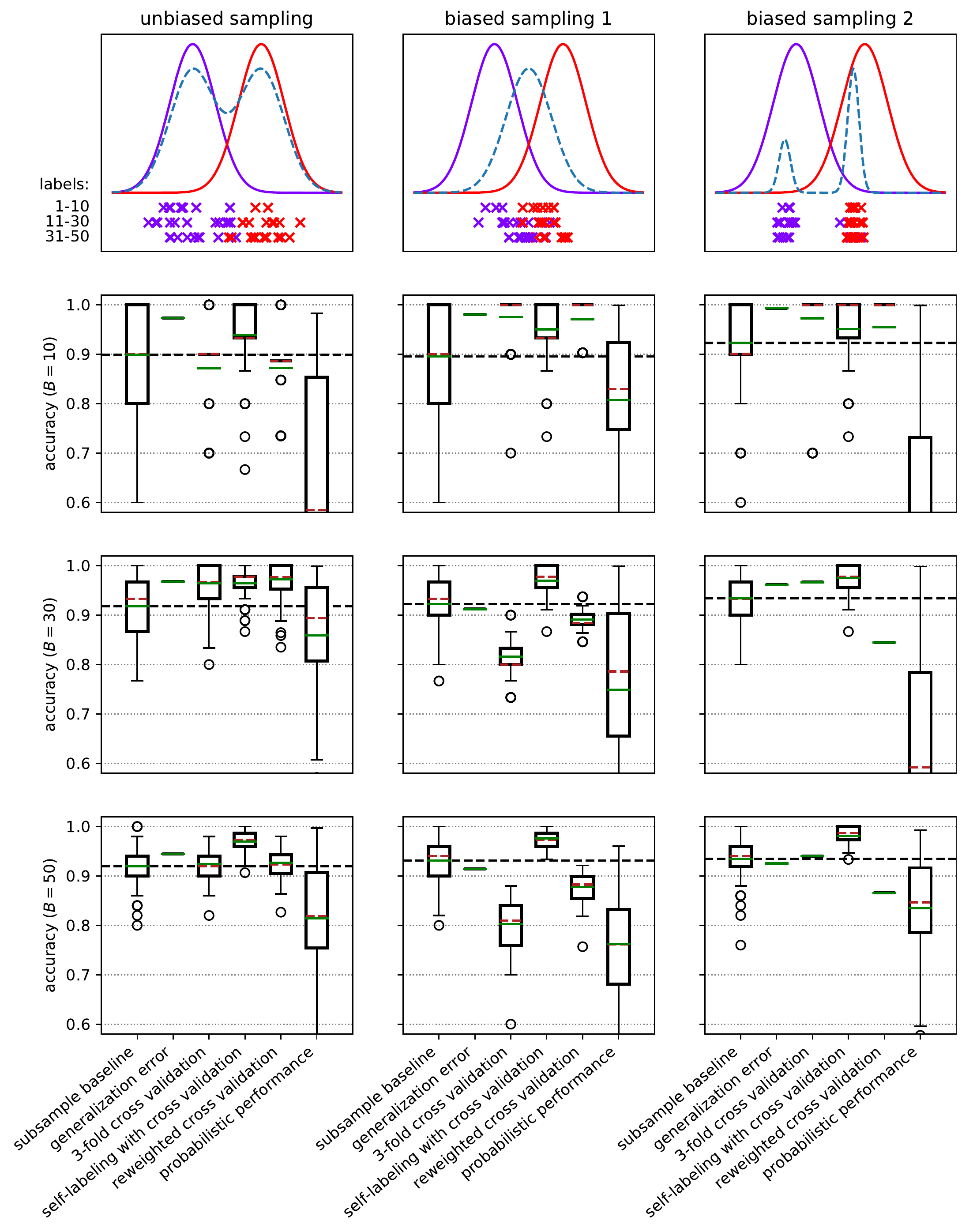}
  \caption{Experimental evaluation of different approaches to assess performance in AL. The upper row shows the dataset and the acquired labels. The main plots show results for different budgets $B$. The boxplots show the mean in green and median in red.}\label{fig:experiment_results}
\end{figure}

All experiments are conducted on the classification task shown in Fig.~\ref{fig:samplingbias_dataset}. In the following, we want to investigate the advantages and drawbacks of the performances estimation algorithms described before. Therefore, we developed three different sampling distributions that can be seen in the upper row of Fig.~\ref{fig:experiment_results} (blue dashed line). These sampling distributions simulate different selection strategies and provide comparability.
Furthermore, the performance is estimated with a varying number of acquired labels. This label budget is $B \in \{10,30,50\}$. Our experiment is a one-shot experiment (real-world like). 
The only labels available for estimation are the ones provided in the upper plots (except for the baselines which have endless labeled data).
Each label is shown as colored crosses indicating the corresponding class.
The upper row ($1-10$) shows the first 10 label acquisitions, the second row acquisitions $11$ to $30$, etc. Hence, a learning task with budget $B=30$ uses the labeled instances from the first two lines.
The true baseline is shown as a horizontal black dashed line. The subsample baseline and all performance estimators are shown using boxplots (median is red, mean is green). 
The classifier is a Parzen window classifier~\cite{chapelle2005active} as used above (the width of the Gaussian kernel is $\sigma$ = 0.2) as this classifier fits the dataset quite well.


\begin{table}[p]
\centering
\scriptsize
\caption{Does a performance estimator address a certain challenge ($+$) or not ($-$)?}
\label{tab:comparison}
\begin{tabular}{lccc}
\hline
                          & \begin{tabular}[c]{@{}c@{}}~~distribution of~~ \\ performance\end{tabular} 
                          & \begin{tabular}[c]{@{}c@{}}~~representative~~\\ training set\end{tabular} 
                          & \begin{tabular}[c]{@{}c@{}}~~sampling~~\\ bias\end{tabular} \\ \hline
Generalization Error      & +                                                                      & --                                                                    & --                                                      \\
3-fold cross validation   & --                                                                     & +                                                                     & --                                                      \\
Self-labeling with CV     & +                                                                      & +                                                                     & +                                                       \\
Reweighted CV             & --                                                                     & +                                                                     & +                                                       \\
Probabilistic Performance & +                                                                      & +                                                                     & +                                                      
\end{tabular}
\end{table}

\subsection{Results and Discussion}

The results in Fig.~\ref{fig:experiment_results} first show that the mean of the subsample baseline is indeed similar to the true baseline. But these baselines are not available in real-world applications as these have lots of labeled data available.

Using this simple dataset with unbiased sampling and $50$ labels, every performance estimate should perform quite well (lower left plot). Understandably, the cross-validation approach performs pretty good as sampling is unbiased. Also, the influence of the reweighting factor is rather small as these values are all similar. The generalization error gets better with the number of labels but we have no information about its confidence. The self-labeling approach overestimates the performance as these self-labeled instances are separable as the classifier's predictions follow a clear decision boundary. The probabilistic approach underestimates the task and is most uncertain (conservative). This is induced by the Beta distribution which adds a uniform prior.

For the first biased sampling (plots in the middle), we would expect underestimation as the sampled dataset is more difficult than the original one. This is perfectly shown by 3-fold cross validation. Neither reweighting nor self-labeling is able to correct this sampling bias. The generalization error seems to fit the problem much better than before, but this is induced by overestimating (like before) a more difficult subsample. Hence, these two errors correct themselves. The probabilistic approach performs similar to the first column.

In the second biased sampling (plots on the right), the selected labeled instances are much simpler than the original dataset as the classifier is able to fit a simple decision boundary. Hence, 3-fold cross validation should overestimate. This is the case with 10 and 30 labels but not with 50 labels. By random 3 purple instances appear in the cluster of red ones. According to the true distributions, this is unlikely but possible. This is the reason that the performance is around $94\%$ which is near the optimum. The other performances show the expected behavior discussed before.






Table~\ref{tab:comparison} summarizes if a strategy implements an approach to the previously proposed challenges. Although two of the tested approaches address all challenges, none of the algorithms is able to provide a reliable performance estimate -- even on such a simple dataset. Hence, we need to focus this problem in AL research as this will make AL systems applicable for more real-world scenarios.

\section{Conclusion and Future Work}

In this article, we gave an comprehensive survey on performance assessment for active learning (AL) at runtime. Thereby, we formalized the problem by defining appropriate baselines and discussing the major challenges:  
 1)~to consider a distribution of performances,
 2)~to have a representative training set when evaluating,
 3)~to correct against sampling bias induced by an intelligent selection strategy.
The evaluation of the identified approaches showed that no method provides reliable performance estimates. We chose a simple dataset and a well-fitting classifier to prove that this problem is one of the most significant ones in AL research: If performance estimation is not possible in simple tasks, there is no reason to try it on real data or complex classifiers.



In the future, we as AL researchers should focus performance estimation for our applications, as this will increase the practical relevance dramatically. We hope this article provides a first direction on the challenges and problems to be solved. The most important but also difficult challenge is the sampling bias correction. Having an unrepresentative set, we are no longer able to estimate performance in regions with few samples. Here, we need models that balance the benefit from active sampling (reducing error) and the benefit from robust performance estimation (reducing uncertainty). Eventually, the best sampling will provide performance estimates with high uncertainty. Hence, performance estimation methods (or stopping criteria) must have an option to actively influence the sampling of labels themselves. One way to performance estimates might be to subdivide the dataspace into several disjunctive regions (urns). Draws from these urns could be interpreted as binomially distributed random variables providing performance distributions. 

Furthermore, insights of estimated performance distributions might lead to new selection strategies optimizing application-oriented demands, for example, the confidence that the accuracy is above 80\% (using quantiles) because the system will pay off if it is above this threshold. 

\subsection*{Acknowledgement}
We thank Adrian Calma, Marek Herde, Christian Gruhl and Sven Tomforde for the insightful discussions and some help with the code.

\newpage
%
%
%

%
\end{document}